# PHA*: Finding the Shortest Path with A* in An Unknown Physical Environment

**Ariel Felner**                                      FELNER@BGUMAIL.BGU.AC.IL
*Department of Information Systems Engineering,*
*Ben-Gurion University of the Negev, Beer-Sheva, 85104, Israel*

**Roni Stern**                                        STERNR2@CS.BIU.AC.IL
**Asaph Ben-Yair**                                    BENYAYA@CS.BIU.AC.IL
**Sarit Kraus**                                       SARIT@CS.BIU.AC.IL
**Nathan Netanyahu**                                  NATHAN@CS.BIU.AC.IL
*Department of Computer Science, Bar-Ilan University*
*Ramat-Gan, Israel, 52900*

## Abstract

We address the problem of finding the shortest path between two points in an unknown real physical environment, where a traveling agent must move around in the environment to explore unknown territory. We introduce the Physical-A* algorithm (PHA*) for solving this problem. PHA* expands all the mandatory nodes that A* would expand and returns the shortest path between the two points. However, due to the physical nature of the problem, the complexity of the algorithm is measured by the traveling effort of the moving agent and not by the number of generated nodes, as in standard A*. PHA* is presented as a two-level algorithm, such that its high level, A*, chooses the next node to be expanded and its low level directs the agent to that node in order to explore it. We present a number of variations for both the high-level and low-level procedures and evaluate their performance theoretically and experimentally. We show that the travel cost of our best variation is fairly close to the optimal travel cost, assuming that the mandatory nodes of A* are known in advance. We then generalize our algorithm to the multi-agent case, where a number of cooperative agents are designed to solve the problem. Specifically, we provide an experimental implementation for such a system. It should be noted that the problem addressed here is not a navigation problem, but rather a problem of finding the shortest path between two points for future usage.

## 1. Introduction

In this paper we address the problem of finding the shortest path between two points in an unknown real physical environment, in which a mobile agent must travel around the environment to explore unknown territories. Search spaces of path-finding problems are commonly represented as graphs, where states associated with the search space are represented by graph nodes, and the transition between states is captured by graph edges. Graphs can represent different environments, such as road maps, games, and communication networks. Moving from one node of the graph to another can be done by applying logical operators that manipulate the current state or by having an actual agent move from one node to another. The sliding-tile puzzle and Rubik's Cube (Korf, 1999) are examples of the first type, while a road map is an example of the second type. Graphs in search problems can be divided into the following three classes:

- **Fully known graphs:** If all the nodes and edges of a graph are stored in the computer, then the graph is fully known. The input for such problems is usually the complete graph which is represented by an adjacency matrix or an adjacency list. A relevant problem in this case would be to find, for example, the shortest path in a road map in which all the nodes and edges are known in advance.

- **Very large graphs:** Graphs that due to storage and time limitations are not completely known and cannot be fully stored in any storage device. Many graphs for search problems have an exponential number of nodes. For example, the 24-tile puzzle problem has $10^{25}$ states and cannot be completely stored on current machines. The input for such problems is usually specified by the general structure of a state in the search space, the different operators, an initial state, and a set of goal states. Only very small portions of such graphs are visited by the search algorithms or are stored in memory.

- **Small, partially known graphs:** The third class contains graphs that represent a partially known physical environment. For example, a mobile agent in an unknown area without a map does not have full knowledge of the environment. Given enough time, however, the agent can fully explore the environment since it is not very large. Due to the partial knowledge, only a small portion of the graph is given as input.

For the class of fully-known graphs, classical algorithms, such as Dijkstra's single-source shortest-path algorithm (Dijkstra, 1959) and the Bellman-Ford algorithm (Bellman, 1958), can be used to find the optimal path between any two nodes. These algorithms assume that each node of the graph can be accessed by the algorithm in constant time. This assumption is valid since all the nodes and edges of the graph are known in advance and are stored in the computer's memory. Thus the time complexity of these algorithms is measured by the number of nodes and edges that they process during the course of the search.

For the second class of graphs the above algorithms are usually not efficient, since the number of nodes in the graph is very large (usually exponential). Also, only a very small portion of the graph is stored in memory at any given time. The A* algorithm (Hart, Nilsson, & Raphael, 1968) and its linear space versions, e.g., IDA* (Korf, 1985) and RBFS (Korf, 1993), are the common methods for finding the shortest paths in large graphs. A* keeps an *open list* of nodes that have been generated but not yet expanded, and chooses from it the most promising node (the *best* node) for expansion. When a node is expanded it is moved from the open list to the closed list, and its neighbors are generated and put in the open list. The search terminates when a goal node is chosen for expansion or when the open list is empty. The cost function of A* is $f(n) = g(n) + h(n)$, where $g(n)$ is the distance traveled from the initial state to $n$, and $h(n)$ is a heuristic estimate of the cost from node $n$ to the goal. If $h(n)$ never overestimates the actual cost from node $n$ to the goal, we say that $h(n)$ is *admissible*. When using an admissible heuristic $h(n)$, A* was proved to be admissible, complete, and optimally effective (Dechter & Pearl, 1985). In other words, with such a heuristic, A* is guaranteed to always return the shortest path. Furthermore, any other algorithm claiming to return the optimal path must expand at least all of the nodes that are expanded by A* given the same heuristic.

An A* expansion cycle is carried out in constant time. This is because it takes a constant amount of time to retrieve a node from the open list, and to generate all its neighbors. The latter involves applying domain-specific operators to the expanded node. Thus the time complexity of A* can also be measured in terms of the number of generated nodes.[1]

In this paper we deal with finding the shortest path in graphs of the third class, i.e., small, partially known graphs which correspond to a real physical environment. Unlike graphs of the other two classes, where a constant number of computer operations are done for node expansion, we cannot assume, for this type of graphs, that visiting a node takes constant time. Many of the nodes and edges of this graph are not known in advance. Therefore, to expand a node that is not known in advance, a mobile agent must first travel to that node in order to explore it and learn about its neighbors. The cost of the search in this case is the cost of moving an agent in a physical environment, i.e., it is proportional to the distance traveled by the agent. An efficient algorithm would therefore minimize the distance traveled by the agent until the optimal path is found. Note that since small graphs are considered here, we can omit the actual computation time and focus only on the travel time of the agent. In this paper we introduce the Physical-A* algorithm (PHA*) for solving this problem. PHA* expands all the mandatory nodes that A* would expand and returns the shortest path between the two points. However, the complexity of the algorithm is measured by the traveling effort of the moving agent. In order to minimize the traveling As will be shown, PHA* is designed to to minimize the traveling effort of the agent by intelligently choosing the next assignment of the traveling agent. As described below, many times the agent chooses to first move to nearby nodes even though they do not immediately contribute to the proceeding of A*.

Unlike ordinary navigation tasks (Cucka, Netanyahu, & Rosenfeld, 1996; Korf, 1990; Stentz, 1994; Shmoulian & Rimon, 1998), the purpose of the agent is not to reach the goal node as soon as possible, but rather explore the graph in such a manner that the shortest path will be retrieved for future usage. On the other hand, our problem is not an ordinary exploration problem (Bender, Fernandez, Ron, Sahai, & Vadhan, 1998), where the entire graph should be explored in order for it to be mapped out. Following are two motivating examples of real world applications for our problem:

- **Example 1:** A division of troops is ordered to reach a specific location. The coordinates of the location are known. Navigating with the entire division through unknown hostile territory until reaching its destination is unreasonable and inefficient. It is common in such a case to have a team of scouts search for the best path for the division to pass through. The scouts explore the terrain and report the best path for the division to move along in order to reach its destination in an efficient manner.

---

1. In fact, for A*, if the open list is stored as a priority queue, then it would take logarithmic time to retrieve the best node. However, for many problems, such as the sliding tile puzzles and Rubik's Cube, a simple first-in first-out queue suffices (Korf, 1993; Taylor & Korf, 1993; Korf, 1997). Likewise, for the linear space versions, such as IDA* and RBFS (which are based on depth-first search), the assumption that it takes constant time per each node is valid. Also we assume that the number of neighbors is bounded.

- **Example 2:** Computer systems connected to networks can be on- or off-line at different times, and their throughput can be seriously degraded due to busy communication channels. Therefore, many networks cannot be represented as fixed, fully known graphs. Transferring large amounts of data (e.g., multimedia files) between two computers over a network can often be time consuming, since the data may be routed through many communication channels and computer systems before reaching their destination. Finding the optimal path between these computer systems could improve the transfer time of large files. Since the network may not be fully known, finding an optimal path between two nodes requires some exploration of the network. An efficient and elegant solution might be to send small packets (operating as scouts) to explore the network and return the optimal path, given that the network is stable for at least a short period of time. Assuming that a computer system on the network is recognized only by its neighboring systems, we are faced with the problem of finding an optimal path in a real physical environment.[2]

In general, it would be worthwhile to search for the optimal path when the following conditions hold:

- Preliminary search (with the usage of scouts) is possible and cheap.

- The optimal path is required for future usage.

Often one might settle for a suboptimal path. However, if the path is needed for a considerable traffic volume, e.g., the path should be traveled a large number of times or the path should be traveled simultaneously by a large number of agents, then finding the optimal path is essential. In this paper we focus on solving such a problem.

The paper is organized as follows. Section 2 provides a more specific formulation of the problem in question. Section 3 discusses related work, and Section 4 presents the PHA* algorithm for a single mobile agent. Several (enhanced) variations are introduced and discussed for this domain, followed by extensive empirical results that demonstrate the superiority of the more enhanced variants pursued. In Section 5 we provide an analysis of PHA* and an overall evaluation of its performance. In Section 6, we provide a number of generalizations for the multi-agent case, where a number of traveling agents are available for solving the problem. Experimental results for these schemes are presented and discussed. Section 7 contains concluding remarks and discusses future research. A preliminary version of this paper appeared earlier (Felner, Stern, & Kraus, 2002).

---

2. Our research is concerned with high-level, abstract graphs and we do not intend to provide here a new applicable routing algorithm. Current routing technologies maintain large databases that store the best paths from node to node, broadcast changes in the network, and update the paths if necessary, thus making essentially the network graph fully known. Also, in some network domains one can create and destroy packages at will and thus does not necessarily have a given number of agents. Our algorithm may be relevant to future network architectures and routing technologies, where routers will not use these databases. This is not far-fetched in view, for example, of the rapid growth of the Internet. It is thus conceivable that in the future storing all the paths would become infeasible.

## 2. Problem Specification

As was mentioned in general terms, the problem is to find the shortest path between two nodes in an unknown undirected graph. More specifically, we assume a weighted graph, where each node is represented by a 2-dimensional coordinate (i.e., its location in the real world), and the weight of an edge is the Euclidean distance between its two nodes. The input to the problem consists of the coordinates of the initial and goal nodes. The other nodes are assumed not to be known in advance. The agent is assumed to be located at the start node. The task is to find the shortest path in the (unknown) graph between the initial node and the goal node for future usage. In order to accomplish that, the agent is required to traverse the graph and explore its relevant parts leading to the desired solution. The agent is allowed to visit nodes and travel from one node to another via existing edges.

We assume here that when a node $v$ is visited by the search agent, the neighboring nodes are discovered, as well as the edges connecting them to $v$. This assumption is not unreasonable, considering, e.g., that (traffic) signs at a road intersection often indicate its neighboring destinations and the lengths of the corresponding road segments that connect it with these locations. Even without road signs, as scouts reach a new location, they can look around, observe the neighboring locations, and assess their distances from their current location. In general, the assumption that the neighboring nodes are discovered instantly is fairly common in search problems and algorithms.[3]

Since the goal of the search is to find the best path to the goal, it is clear – given an admissible heuristic – that the agent must expand all nodes expanded by A*, as A* is optimally effective (Dechter & Pearl, 1985). Let $C$ be the length of the shortest path from the initial node to the goal node. A* will expand all the nodes, such that, $f(n) = g(n) + h(n) < C$ and some of the nodes for which $f(n) = C$. We will refer to these nodes as the (set of mandatory) *A* nodes*. As stated above, the agent must visit all the A* nodes in order to find the shortest path. However, it may need to visit additional nodes.

We make the following fundamental observations with respect to the problem in question:

- First, even if the set of A* nodes is known in advance, the agent may need to visit additional nodes while traversing related portions of the graph. This is because the shortest path between two of the A* nodes may include graph nodes that do not belong to the A* nodes, i.e., their $f$ value is greater than $C$. Given the A* nodes, finding the shortest path that visits all of them – this should not be confused with the shortest path between the origin node and the goal node – could be considered as solving the traveling salesman problem (TSP) with respect to the set of A* nodes. Note that the TSP solution may include nodes that do not belong to the A* nodes.

- Second, the agent does not know the A* nodes in advance. These nodes are added to the open list and they are expanded as the search progresses. Thus our agent cannot use a solution to the TSP, since TSP assumes that the nodes to be visited are provided as input.

---

3. There are, however, domains where this assumption may not hold. In such domains, a node becomes fully known only when the agent reaches it physically. In this work we restrict ourselves to the above assumption. Other domains will be addressed as part of future work.

In most of the cases, the order in which A* nodes are expanded is very different from the order in which they are visited according to the TSP solution. Thus the minimal path traversing the A* nodes cannot be used.

- Third, when a node is added to the open list the agent cannot know whether or not it belongs to the A* nodes, since $C$ is known only after the search is concluded. Consider a node $n$ that is in the open list, but is not at the head of the open list. Suppose further that the agent is physically located near that node. It should decide whether to slightly extend its path and visit node $n$ or skip $n$ and continue to the node at the head of the open list. If $n$ turned out to belong to the A* nodes, then visiting it now may prove very beneficial. (This is because $n$ might reach the head of the open list when the agent will be physically located far away from it, so that visiting $n$ at that point will incur a significant travel cost.) However, if it turns out that $n$ does not belong to the A* nodes, then the (small) detour of visiting it has proven useless. Intuitively, however, a decision never to visit such would result in a very bad strategy. Thus the agent may visit nodes that do not belong to the A* nodes because of future expected benefits. The actual decision as to whether or not to visit $n$ will depend on the distance between the agent's location and $n$ (at the time of the decision) and on the agent's estimate as to whether $n$ belongs to the set of A* nodes.

In the following sections we present the PHA* algorithm for efficient exploration of a graph, in order to find the shortest path between two given nodes by a single traveling agent, as well as by multiple agents. We study different heuristics that direct the agent to make an intelligent decision, in an attempt to achieve a small overall travel cost.

## 3. Related Work

Much research has been devoted to guiding a mobile agent for exploring new and unknown environments in order to study them and map them out. Our work is different, in the sense that it explores merely the necessary regions of the graph in order to retrieve the shortest path between two nodes and not the entire graph. Most of the literature in this area deals with a physical mobile robot that moves in a real environment. The published research focuses usually on the issue of assisting the robot to recognize physical objects in its environment. We refer the reader to (Bender et al., 1998), which contains an extensive survey of various related approaches and state of the art techniques.

Another class of algorithms is navigation algorithms. A navigation problem is concerned with navigating a mobile agent to the goal as fast as possible, not necessarily via the shortest (optimal) path. A navigator will always proceed towards the goal, ignoring whether the trail traversed thus far lies on the shortest path. Deviations from the optimal path are neglected since the navigation problem is reconsidered after every move with respect to a new source node, i.e., the current position of the agent. A navigation algorithm halts when the mobile agent reaches the goal. The path passed usually lacks importance and is usually not optimal. Our problem, on the other hand, is to find an *optimal* path to the goal node for future usage. Even if the agent finds a path to the goal node, the search continues until the shortest path to the goal is found. Next, we describe briefly some of the work done on navigation in partially known graphs.

(Cucka et al., 1996) have introduced navigation algorithms for sensory-based environments such as automated robots moving in a room. They have used depth first search (DFS)-based navigation algorithms, that use a heuristic function for choosing the next node that the agent should go to.

Real-Time-A* (RTA*) (Korf, 1990) and its more sophisticated version, Learning Real-Time-A* (LRTA*), are also algorithms for finding paths between two nodes in a graph. However, they deal with large graphs and assume that there is a constraint on the time of computation and that a move should be retrieve in a given constant time. Thus a limited search is performed, and the node with best cost in the search frontier is picked. The problem solver then moves one step along the path to that node. The search then continues from the new state of the problem solver. The merit of node $n$ (in RTA* and LRTA*) is $f(n) = g(n) + h(n)$, similarly to A*. Unlike A*, though, $g(n)$ is the actual distance of node $n$ from the current state of the problem solver, rather than from the original initial state. The difference between RTA* and LRTA* is that after the search is terminated, LRTA* also stores the heuristic estimation value of each node visited by the problem solver. Also the method that successor nodes are chosen are different for the two variations. Korf (Korf, 1990) proves that over a large number of runs, where for each run the start node is selected at random, the stored value of each node visited by the LRTA* problem solver converges to the optimal distance to the goal. Both RTA* and LRTA* are significantly different from our approach, as they assume that a node can be expanded in the computer's memory without an agent having to physically visit that node. (Also, these algorithms are designed for large graphs.) Furthermore, RTA* does not find the optimal path to the goal. A trivial version of LRTA* could be used to solve our problem, e.g., by limiting the search depth to one level, so that every node visited by the agent could be physically expanded. However, such a variant will not be competitive with our approach, as it will perform like a simple hill-climbing procedure. In addition, in order to attain the optimal path, LRTA* has to select many start nodes at random. This is not relevant in our case, as we are given only one initial node.

MARTA* (Knight, 1993) is a multi-agent version of RTA*. In MARTA* every agent runs RTA* independently. Kitamura et al. (Kitamura, Teranishi, & Tatsumi, 1996) have modified MARTA* by using coordination strategies based on attraction and repulsion. These strategies are employed only in tie-breaking situations. When using a repulsion strategy, the idea is to spread agents, such that each agent intends to maximize its distance from the others. Again, the path provided by this algorithm is not optimal and also, agents do not need to physically visit a node in order to expand it. This work has inspired the algorithms presented in this paper, as far as handling our multi-agent mutual decision is concerned.

Life-long planing A* (LPA*) (Koenig & Likhachev, 2002b) is a remarkable algorithm that generalizes A* to handle a dynamically changing graph. LPA* is activated every time the graph was changed in order to find the current shortest path from the same given start and goal nodes. It utilizes the fact that much of the old data explored by previous runs of LPA* are still valid in the current run. A* is a special case of LPA* where the entire graph has not been explored yet.

D*-lite (Koenig & Likhachev, 2002a) applies LPA* to the case that a mobile robot needs to find the shortest path in an unknown environment or in an environment that changes dynamically (i.e., where edges are added and deleted at all times). In LPA*, the start node is identical for all the runs. In D*-lite, however, the robot moves along the path and calculates a new shortest path from its current location. D*-lite modifies LPA* so that old data from previous runs will be efficiently used in the case that the start node is also changed according to the new location of the robot. D*-Lite is actually a simplified version of a previous algorithm D* by Stenz (Stentz, 1994).

The main difference between these algorithms and our approach is that they, too, expand a node in the computer's memory without requiring that the mobile agent physically visit that node. Indeed, following every move of the robot in D* Lite, changes in the graph are provided immediately ; the robot does not need to physically visit nodes in order to gather firsthand this information. The task of the agent, in the context of D*, is to repeatedly determine the shortest path between the current location of the robot and the goal location as the edge costs of the graph changes while the robot moves. D* lite does not find a path and returns it. It is simply a navigation algorithm that guides the agent to the goal node based on previous and new information about the terrain.

An agent operating in the real world must often choose between maximizing its expected utility (according to its current knowledge of the "world") and learning more about its environment, in an attempt to improve its future gains. This problem is known as the trade-off between exploitation and exploration in *reinforcement learning* (Kaelbling & Moore, 1996). Argamon et al. (Argamon-Engelson, Kraus, & Sina, 1998, 1999) address the trade-off between exploration and exploitation for an agent that moves repeatedly between two locations. They propose a utility-based on-line exploration algorithm which takes into account both the cost of attempting to improve the currently best route known and an estimate of the potential benefits over future task repetitions. If the expected utility from exploration is positive, then the agent takes actions to improve its route; otherwise, it continues using the known path. The authors compare the utility-based on-line exploration with a heuristic backtracking search algorithm that exhaustively searches the graph before starting to perform the task, and with a randomized interleaved exploration algorithm. They assume that the agent knows a path between any two nodes, while we make no such assumption.

Argamon et al. also suggest that the larger the number of times that the task is repeated, the more the merit of interleaved exploration diminishes. If the agent is required to move back and forth between two nodes a large number of times, there is no need to decide on-line whether to exploit or explore; instead, the shortest path should be found as soon as possible. Thus a good search algorithm may prove useful. In this respect our work complements Argamon et al., as it provides efficient search algorithms in situations where the optimal path is needed in advance. In contrast, applying the techniques of Argamon et al. in these situations yields poor results as demonstrated in their experiments.

Roadmap-A* (Shmoulian & Rimon, 1998) is a more sophisticated single agent navigation algorithm. It chooses to navigate to a node that is assumed to be close to the goal node. The algorithm is supervised by a high-level procedure called $A_\varepsilon^*$ (Pearl & Kim, 1982). Instead of always selecting the best node from the open list, $A_\varepsilon^*$ allows the search agent to choose from a set of "good nodes". This set is called the *focal set*. The *focal* is a set of nodes from the open list whose $f$ value is greater than the value of the best node by no more than $\varepsilon$. Once the focal nodes are determined, a local search is performed to navigate the agent to one of these nodes, which is believed to be close to the goal node. The role of the high-level phase is to prevent the navigating agent from going in the wrong direction by considering also the path traveled thus far.

In Roadmap-A\*, $\varepsilon$ is a pre-specified constant, which determines the trade-off between the local search and A\*. For example, $A_0$ is A\* while $A_\infty$ is just a local search, choosing at each iteration any node that is believed to be close to the goal node. This algorithm halts when the goal node is reached, and thus for $\varepsilon > 0$ the optimal path might not be known. The paradigm of Roadmap-A\* is similar to ours, in the sense that a node is known only after the agent explores it. In fact, in the trivial case where $\varepsilon = 0$, Roadmap-A\* is very similar to our approach with the simple heuristic "shortest-known path" (presented in Subsection 4.1 below). Further comments as to the basic difference between RoadmapA\* and PHA\* are provided in Section 5.

In summary, most of the above listed algorithms are navigation algorithms, i.e., they do not necessarily require an agent to physically visit a node in order to expand it, and do not necessarily return the optimal path to the goal node. Thus they inherently solve a different problem from the one pursued in this paper.

## 4. PHA\* for a Single Agent

We now turn to the description of the PHA\* algorithm, focusing first on the case where only a single mobile agent is available.

Nodes in the environment can be divided into *explored* and *unexplored* nodes. Exploring a node means physically visiting that node by the agent, and learning about its location and the location of its neighbors. Our new algorithm PHA\* activates essentially A\* on the environment. However, in order to expand a node by A\*, this node must first be explored by the agent in order to obtain the relevant data associated with it (i.e., neighboring nodes and incident edges). Throughout the discussion in this paper we treat PHA\* as a two-level algorithm. Although in principle PHA\* could also be viewed as a one-level algorithm (see further discussion in Subsection 4.2), we find its two-level presentation to be more well-structured and better understood conceptually. The two-level framework consists of a high-level and a low-level routine. The high level (which invokes the low level at various stages of PHA\*), acts essentially like a regular A\* search algorithm. It chooses at each cycle a node from the open list for expansion. The heuristic function $h(n)$ used here is the Euclidean distance between $n$ and the goal node. (This heuristic is admissible of course, by definition.) If the node chosen by the high level has not been explored by the agent, the low level, which is a navigation algorithm, is activated to navigate the agent to that node and explore it. After a node has been explored by the low level it is expandable by the high level. If the chosen node has already been explored, or if its neighbors are already known, then it is readily expandable by the high level without the need to send the agent to visit that node. The pseudo-code for the high level is given below.

```
high-level(open-list) {
.   while(open-list is not empty) {
.       target = best node from open-list;
.       if target is unexplored then {
.           explore(target) by the low level;
.       }
.       expand(target);
.   }
}
```

## 4.1 Low-Level Algorithms

The high-level algorithm, A*, chooses to expand the node with the smallest $f$ value in the open list, regardless of whether or not the agent has already visited that node. If the chosen node has not been visited by the agent, the low level instructs the agent to visit that node. We call this node the *target* node for the low level. In order to reach the target node, we must use some navigation algorithm. We have implemented a number of navigation variants for the low level. We first describe simple algorithms which only use known information about the graph. We then present more efficient algorithms, which also explore the graph during the navigation and provide new information for the high level. We assume that the agent is in the *current* node and that it needs to navigate to the *target* node.

### 4.1.1 SIMPLE NAVIGATION ALGORITHMS

- **Tree path:** Like every best-first search, A* spans the nodes which it generates in a tree which is called the *search tree*. Every known node is a node in the search tree. The most trivial way to move from one node to the other is through the search tree. The tree-path algorithm instructs the agent to move from the current node to the target node through the shortest path between them in the search tree. In other words, the agent will walk up the tree from the current node until it reaches an ancestor of the target node, and then walk from that node to the target node. This is a trivial algorithm, and is presented here mainly for comparison purposes.

- **Shortest known path:** Some of the nodes of the search tree have already been explored by the agent, so that all of their incident edges are known. The search tree nodes plus the additional edges of the explored nodes can be viewed as a subgraph that is fully known. All the nodes of this subgraph are connected because they are all part of the search tree. Using this subgraph, we can calculate the shortest path to the target node via known nodes and edges. As mentioned above, finding the shortest path in a known graph can be done easily, so the agent simply computes this shortest path to the target node and travels along that path.[4]

- **Aerial path:** Assuming that the agent is able to move freely in the environment and is not restricted to the edges of the graph, we can simply move the agent from the current node to the target node via the straight line connecting these nodes. This method may be relevant when the search agents are highly mobile, and they explore the environment for agents that are restricted to travel only along the edges. Note that the length due to "aerial path" can never be greater than the length due "shortest known path".

---

4. This navigation algorithm is similar to the local A* search in Roadmap-A* for the trivial case where $\varepsilon = 0$. In Roadmap-A*, the shortest path to the target node is determined in the known graph and the agent moves along that path.

### 4.1.2 DFS-BASED NAVIGATION ALGORITHMS

In the simple navigation algorithms described above, the exploration of new nodes is done only by the high-level algorithm. Thus the low level does not add any new knowledge about the graph, and in that sense it is inefficient. We propose here more intelligent navigation approaches for finding a path to the target that can pass also trough unexplored nodes. These approaches provide the following advantages: The paths that are currently known to the agent may be much longer than other paths that have not been explored yet. It may prove more efficient to navigate through unknown parts of the graph if they seem to lead to a better path to the target. A more important advantage is that while navigating through unknown parts of the graph, the agent might visit new nodes that have not been explored and explore them on the fly. This may save the need to travel back to those nodes at a later time, should they be selected for expansion by the high-level algorithm.

The above advantages suggest the use of a DFS-based navigation for the low level. In a DFS-based navigation algorithm, the search agent moves to a neighboring node, that has not been visited, in a typical DFS manner. The algorithm backtracks upon reaching a dead-end and the search continues until it reaches the target. If there is more than one neighbor, we use a heuristic to evaluate which neighbor is more likely to lead faster to the target, and visit that node first. We have experimented with the following DFS-based navigation algorithms that were proposed by (Cucka et al., 1996):

- **Positional DFS (P-DFS):** This DFS-based navigation algorithm sorts the neighbors according to their Euclidean distance from the target node, choosing the node with minimum distance to the target node first.

- **Directional DFS (D-DFS):** This DFS-based navigation algorithm sorts the neighbors according to the direction of the edges between them and the current node $v$. It first chooses the node $u$ for which the difference in angle between the line segments $(v, u)$ and $(v, t)$ is the smallest, where $t$ denotes the target node. In other words, the nodes are prioritized by the directional difference between them and the target node, giving priority to nodes that differ the least.

- **A*DFS:** A*DFS is an improved version of P-DFS. At each step the agent chooses the neighbor $w$ that minimizes the sum of the distances from the current node $v$ to $w$ and from $w$ to the target node $t$. We call it A*DFS since it uses a cost function which is similar to that of A*, i.e., $f(n) = g(n) + h(n)$.[5] Note, however, that this cost function is used here locally to find a path from the current node to the target node.

---

5. A generalized version of navigating with a cost function similar to that of A* called "robotic A*" (RA*), was also proposed by (Cucka et al., 1996); the node $w$ is either a neighbor (of $v$) or an already visited node.

This is different from the high-level A* which uses this cost function to find a path from the input initial state to the input goal state.

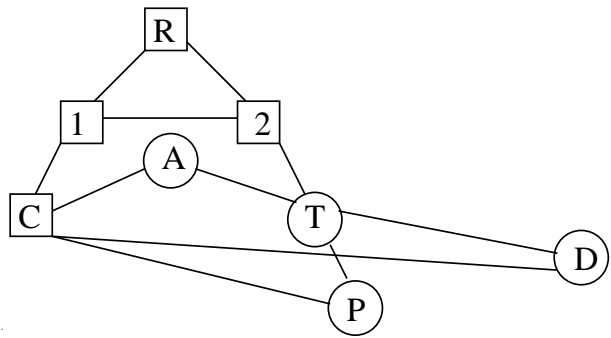

Figure 1: Illustration of various low-level navigation algorithms.

Figure 1 illustrates the navigation algorithms listed above. Let $R$ denote the source node, and suppose that the search agent is currently at node $C$, and that the high-level procedure chooses to expand node $T$. The squared nodes have already been visited by the agent, i.e., they have already been explored. These nodes and the edges connecting them comprise the tree spanned by the high-level A* search. Since $T$ has yet to be explored, the low-level procedure will now navigate to the target node $T$. The tree path will navigate along the path $C - 1 - R - 2 - T$, whereas the shortest known path will navigate along the path $C - 1 - 2 - T$. Note that since node $A$ has yet to be explored, the path from $C$ to $T$ via $A$ is not known at this point. The aerial path will go directly from $C$ to $T$. Using one of the DFS-based navigations, the agent will move to $T$ via $P$, $D$, or $A$ depending, respectively, on whether P-DFS, D-DFS, or A*DFS was used. The benefit of the DFS-based algorithms is that they explore new nodes during the navigation (nodes $P$, $D$, and $A$ in the above example), and they will not revisit such nodes, should the high-level procedure expand them at a later stage.

## 4.2 Enhanced PHA*

### 4.2.1 PHA* AS A ONE-LEVEL PROCEDURE

As mentioned in the previous subsection, PHA* can be presented in principle as a one-level algorithm. This can be done as follows. Whenever the best node in the open list is known (i.e., it has been explored), an expansion cycle of A* takes place in the background, and a new best node is determined. Upon arriving at a node, the agent makes a navigation decision as follows:

- If the best node in the open list is one of the current node's neighbors, then the agent moves to that node.

- Otherwise, the agent moves to the neighboring node that minimizes the relevant heuristic function (among the variants proposed in the previous subsection).

Any of the these heuristics would be valid for a heuristic function of this one-level algorithm. (The latter should not be confused with the heuristic function that is associated with the A* expansion cycle.) For example, if the agent is at node $v$, then using A*DFS it will visit the neighbor $w$ that minimizes the sum of the distances from the current node $v$ to $w$ and from $w$ to the best current node in the open list.

The compact one-level presentation notwithstanding, we prefer – for reasons of clarity – to use the two-level formulation of PHA*. We believe that the clear distinction between the high-level A* and the low-level navigation procedure provides an overall framework that is well-structured and conceptually more clearly understood. In addition, the two-level framework lends itself naturally to the two enhancements presented in the following subsections.

These enhancements draw on the basic principle that navigation might proceed not necessarily to the best node, but to a different node that is fairly close to the current location of the agent. (The idea being that in the long run this would prove beneficial.) This principle can be realized under two main scenarios: (1) When navigating to the best node, the agent might choose first to visit a nearby neighbor, and (2) the procedure might choose to ignore the best node in the open list and select instead a different node from the open list which is very close to the agent's location. In the context of the two-level framework, the first scenario corresponds to a low-level enhancement (see I-A*DFS below), and the second scenario corresponds to a high-level enhancement (see WinA*, subsection 4.2.3).

For all of the above reasons, we choose to stick with our proposed two-level approach of PHA*.

### 4.2.2 IMPROVED LOW LEVEL: I-A*DFS

The DFS-based navigation algorithms explore new nodes as they traverse the graph, thereby avoiding future navigations should these nodes be selected later for expansion by the high level. While this is very beneficial, as can be seen from the experimental results of the next subsection, we can take this approach much further.

Suppose that the agent is navigating to a target node. Along the way, it may pass near nodes that have a small $f$ value without visiting them, as they are not on the path to the target node according to the navigation algorithm. This is counter-productive, since nodes with small $f$ values are likely to be chosen for expansion by the high level in the near future. Visiting such nodes when the agent is nearby, may save a lot of traveling effort in the future. In order to motivate the agent to visit such nodes, we want to identify them and artificially decrease their cost value (without changing the value of other nodes).

To incorporate this notion, we introduce the Improved A*DFS (I-A*DFS) variant. The basic concept is that while navigating to a target, the low level will select the next node to visit by considering not only its approximate distance from the target but also the node's $f$ value. On its way to the target, I-A*DFS should tend to visit, on the one hand, nodes with a small $f$ value, and avoid visiting, on the other hand, nodes that are completely off track.

Let $T$ and $n$ denote, respectively, the target node and the neighboring node that is being currently evaluated. Also, let $f(.)$ denote the $f$ value of a node provided by the high-level A*, and let $c_1$, $c_2$ denote constants to be specified. We used the following heuristic function for selecting the next node by I-A*DFS:

$$h(n) = \begin{cases} \text{A*DFS}(n) \cdot \left(1 - c_1 \left(\frac{f(T)}{f(n)}\right)^{c_2}\right) & \text{if } n \in \text{OPEN} \\ \text{A*DFS}(n) & \text{otherwise.} \end{cases} \tag{1}$$

If a neighbor $n$ is not in the open list, then its $h(n)$ value due to A*DFS remains intact. If, however, the neighboring node is in the open list, then I-A*DFS considers also the goodness of its $f$ value. The node's $h(n)$ is adjusted according to a product term that decreases with the node's $f$ value (i.e., a node with a small $f$ value will be assigned a smaller heuristic)[6]. Specifically, the goodness of $f$ is measured by the ratio $f(T)/f(n)$. The target node $T$ has the smallest $f$ value among the nodes in the open list (for otherwise it would not have been selected for expansion by the high level) and therefore $0 < f(T)/f(n) < 1$. If $f(T)/f(n)$ is close to 1, then $f(n)$ is close to $f(T)$. In this case, it is highly probable that node $n$ will be visited by A* in the next few steps. Thus we want to assign a higher priority to such a node to be visited by the agent, by decreasing its heuristic value. If, however, $f(n) >> f(T)$ (i.e., $f(T)/f(n) \to 0$), then it is highly unlikely that node $n$ will be selected anytime soon by the high level A*. It is of no interest to raise the node's priority, in such a case, and its A*DFS heuristic should be retained, just like other nodes that are not in the open list.

The expression provided in (1) meets all of the above requirements. If $f(n) \approx f(T)$, then the term $1 - f(T)/f(n)$ becomes small, and the overall $h$ value for such a node decreases. This provides the agent with an option to visit nodes that are in the open list and which have small $f$ values, even though their A*DFS heuristic is not the best. If, on the other hand, $f(n) >> f(T)$, then the term $1 - f(T)/f(n)$ will approach 1, having a negligible effect on $h(n)$. The main reason for multiplying the A*DFS heuristic by $1 - f(T)/f(n)$ (and not by $f(n)/f(T)$, for example) is to leave intact the cost value of a node with a relatively large $f$ value, so that it can continue to compete (in a local heuristic sense) with nodes which are not in the open list. The free parameters, $c_1$ and $c_2$, do not affect qualitatively the performance of I-A*DFS, but merely add to the module's overall flexibility.

We have experimented with various constants for $c_1$ and $c_2$, in an attempt to determine optimal performance. Our extensive empirical studies have shown that $c_1 = 0.25$ and $c_2 = 2.5$ produced the best performance. Our experiments have also demonstrated that using I-A*DFS yielded better results than those obtained by the other navigation algorithms listed in Subsection 4.1.2.

Figure 2 illustrates the difference between A*DFS and I-A*DFS. The numeric values of the nodes indicate the order by which they are expanded by A*. Suppose that the agent is currently located at node $C$ and that node 1 is the target. A*DFS will navigate to the target via node 5, since this node has the best $f(= g + h)$ value for the scenario described. Once at node 1, the agent will have to travel back to the other side of the graph, as node 2 is selected (by the high level) to be expanded next. The agent will then go back to node 3 and eventually reach the goal via node 4. I-A*DFS, on the other hand, will navigate from $C$ to node 1 via node 2; although node 2 is not assumed to be on the shortest path to node 1, it has a smaller $f$ value than node 5. Thus I-A*DFS chooses to visit node 2 first. Incorporating this principle saves a considerable amount of travel cost. When the agent will be located at node 1 and the next node to be expanded will be node 2, the high level will expand it immediately, as it has been explored before by the agent, and will thus be readily available. Thus the agent will travel directly from node **1** to node **3** and will avoid navigating back and forth between opposite sides of the graph.

---

6. Since A*DFS(.) and $f(.)$ measure distances on the graph, they represent, essentially, the same scale. Thus they can be combined directly.

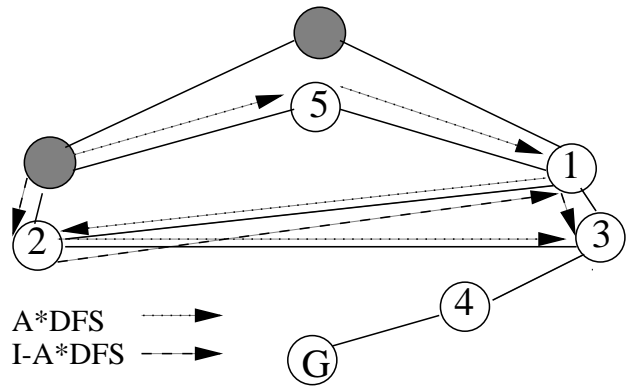

Figure 2: An example of A*DFS versus I-A*DFS navigation.

### 4.2.3 Improved High-Level: WinA*

A* expands the nodes from the open list in a best-first order according to their $f$ value. This order is optimal when the complexity of expanding a node is $O(1)$. However, in a real physical environment, where node expansion requires an agent to perform costly tasks, it is not always efficient to expand the current best node. Consider, for example, a nearby node that is not the best node in the open list, but whose $f$ value is sufficiently small, such that with high probability it would be selected for expansion by A* in the next few iterations. An intelligent agent will choose to explore such a node first, even though it is not currently the best node in the open list.

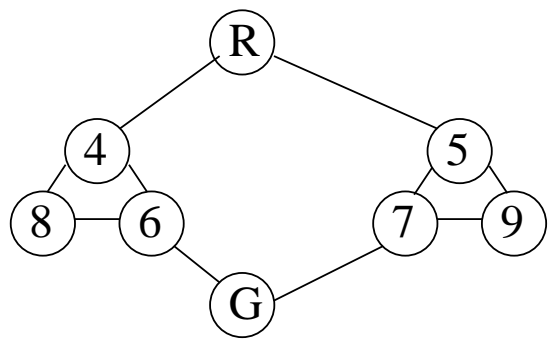

Figure 3: An example illustrating the disadvantage of A*.

The above principle is illustrated, for example, by the subgraph of Figure **3** which contains two node clusters. The numeric label of each node is associated with its $f$ value. An agent visiting the nodes in a best-first order (i.e., the order by which A* expands them), will have to travel back and forth from one cluster to the other. A much better approach would be to explore all the nodes in one cluster and then move to the other cluster, thereby traveling only once from one cluster to the other.

In order to incorporate this capability into our algorithm, we generalized A* to what we call Window A* (WinA*). While A* chooses to expand the node with the lowest $f$ value, WinA* creates a set (i.e., a window) of $k$ nodes with the smallest $f$ values and then chooses one node from the set for expansion[7]. Our window uses the same principle of $A_\varepsilon^*$ (Pearl & Kim, 1982) which was mentioned before. After constructing the window we select from it a node for expansion. Our objective is to minimize the traveling effort of the agent, and not to reduce, necessarily, the number of expanded nodes. Thus rather than selecting only those nodes that have a small $f$ value, we choose also nodes that are sufficiently close to the location of the agent. Having experimented with a large number of combinations, we concluded that the best way of capturing these two aspects was by simply taking their product. Thus we order the nodes of the window by the cost function

$$c(n) = f(n) \cdot \text{dist}(\text{curr}, n),$$

where $n$ is the node evaluated, $f(n)$ is its $f$ value, and $\text{dist}(\text{curr}, n)$ is the distance between $n$ and the current location of the agent. We choose to expand the node with the smallest cost $c$. (It is sensible to combine $f(n)$ and $\text{dist}(\text{curr}, n)$ in the above manner, as both are expressed in the same distance units.) Note that if a node with a small $f$ value is not chosen for expansion, then its $f$ value relative to other nodes in the open list will tend to decrease over time. This is because the $f$ value of newly generated nodes is monotonically increasing, as the heuristic used is consistent and admissible. This property reduces the chance for starvation. (At least we have not encountered this phenomenon in our experiments.)

Our intention was to demonstrate that combining these two factors, in a manner that favors nearby nodes having a small f value, indeed yields enhanced performance. We have tried many functions that combine the two factors (e.g. weighted sum) but choose in this paper to only discuss the product, $c(n) = f(n) \cdot \text{dist}(\text{curr}, n)$, since it provided the best results.

Combining this modified high-level variant with the low-level navigation creates some technical difficulties, due to the fact that we no longer expand nodes from the open list in a best-first order. Recall that standard A* expands a node by generating its neighbors and putting the node in the closed list. When a node $v$ is in the closed list, the shortest path from the source node to $v$ is known. Hence, when the goal is expanded we have found the shortest path to it, and the search can terminate. However, in WinA* a node may be expanded although there exists another node with a smaller $f$ value that has not been expanded yet. In other words, when a node $v$ is expanded, it does not necessarily imply that the best path to $v$ has been found. Expanding a node with a smaller $f$ value might discover a better path. Thus the search cannot simply terminate once the goal node is chosen for expansion.

This problem is solved by splitting the standard node expansion stage into two phases:

---

7. In a related algorithm that we have derived, $k$-best first search (KBFS) (Felner, Kraus, & Korf, 2003), a window of size $k$ is determined from the open list, and all of the window nodes are expanded at the same stage. The neighbors of all the nodes are generated and added to the open list, and only then does a new iteration begins.

1. **Node expansion.** Expanding a node means visiting the node, generating all its neighbors, and adding them to the open list. This stage takes place immediately for each node chosen by the high level.

2. **Node closing.** Closing a node means removing it from the open list and putting it on the closed list. This takes place only after all the nodes with a smaller $f$ value have been explored. This ensures, essentially, that a node will be placed in the closed list only when the best path to it from the source node has been found (See Section 5 for further comments). Thus the search will continue, even if the goal node has been expanded, until it is placed in the closed list. Only when the goal node is placed in the closed list, does the search terminate.

Following is the pseudo-code for WinA*. Note that the standard expansion is divided according to the above two phases. At the end of each cycle, the algorithm attempts to close as many nodes as possible.

```
WinA*() {
.    while (goal is not in closed-list) {
.        target = node from window that minimizes f(node)·dist(current, node);
.        if target is unexplored then
.            explore(target) by low level;
.        expand(target);
.        while (best node (with minimal f value) in open-list was expanded)
.            close(best node);
.    }
}
```

## 4.3 Experimental Results

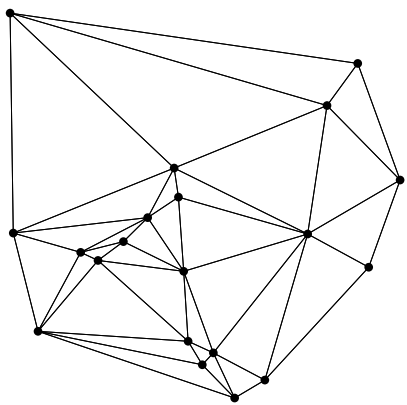

Figure 4: A 20-node Delaunay graph.

We have experimented with Delaunay graphs (Okabe, Boots, & Sugihara, 1992), which are derived from Delaunay triangulations. The latter are computed over a set of planar point patterns, generated by a *Poisson point process* (Okabe et al., 1992). Points are distributed at random over a unit square, using a uniform probability density function. A Delaunay triangulation of a planar point pattern is constructed by creating a line segment between each pair of points $(u, v)$, such that there exists a circle passing through $u$ and $v$ that encloses no other point. Such a triangulation can be characterized, in a sense, as one where each point is joined by a line segment to each of its nearest neighbors but not to other points. (We will refer to this type of Delaunay graphs as *regular* Delaunay graphs.) We have used the Qhull software package (Barber, Dobkin, & Huhdanpaa, 1993) to construct Delaunay triangulations (i.e., Delaunay graphs) over sets of points that were generated at random in a unit square. Figure 4 illustrates a 20-node Delaunay graph.

In principle, the characteristic whereby each node is connected to all its neighbors seems suitable for representing real road maps, which are the main object of our research. In practice, however, additional characteristics should be accommodated to capture more adequately a real road map. Thus we have also pursued *sparse* and *dense* Delaunay graphs that can be obtained from regular Delaunay graphs by random deletion and addition of edges, respectively. (See Appendix A for a more detailed discussion.)

### 4.3.1 LOW LEVEL EXPERIMENTAL RESULTS

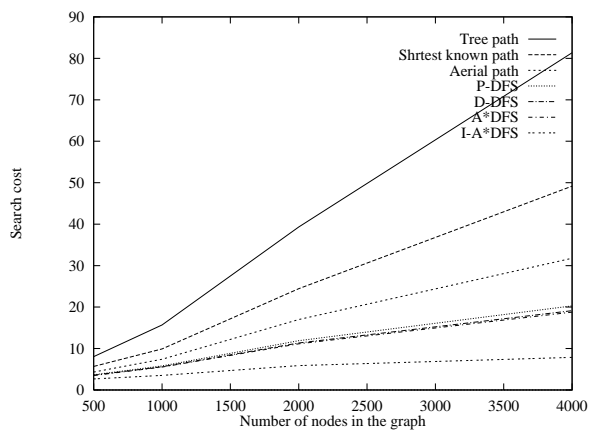

Figure 5: Search cost versus the number of nodes of regular Delaunay graphs for various low-level algorithms.

Figure 5 displays the traveling distance (or search cost) of the agent as a function of the number of nodes in the Delaunay graph (i.e., 500, 1000, 2000, and 4000 nodes). The graphs depicted correspond to the various low-level algorithms that **PHA\*** was tested on. Every data point (here and in all the other experiments) corresponds to an average of 250 different pairs of initial and goal nodes, that were picked at random. The average optimal path observed was about 0.55.[8] The figure clearly demonstrates the higher efficiency of the more involved algorithms. In particular, I-A\*DFS is consistently superior to all the other algorithms for all graph sizes. For a graph of size 4000, for example, it outperformed the most simple algorithm by a factor of more than 10, and outperformed the basic A*DFS by a factor of more than 2. Note that the search cost increases as the number of nodes grows, i.e., as the domain becomes denser or more connected. This is attributed to the fact that as the number of nodes grows, so does the number of nodes in the closed list that the I-A*DFS procedure has to visit.

---

8. Note the closeness between the average optimal path observed (i.e., 0.55) and the expected arc length of a random graph defined over the same set of points (i.e., 0.521) (Ghosh, 1951).

The relative performance of the various algorithms we have considered remained the same for sparse and dense Delaunay graphs (see Appendix A).

### 4.3.2 EXPERIMENTAL RESULTS FOR WINA*

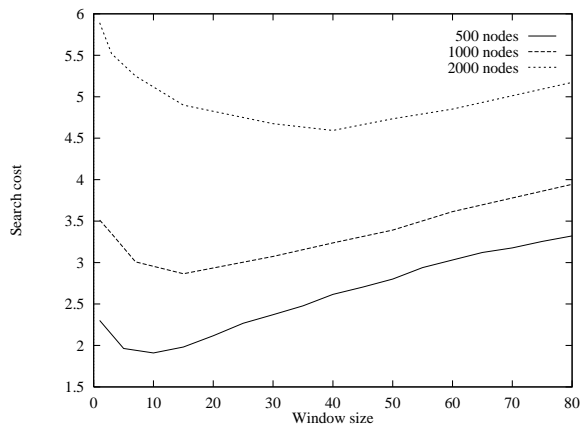

Figure 6: Search cost of WinA* versus window size for various sizes of regular Delaunay graphs.

Our experiments show that using WinA* as the high-level procedure of PHA* leads to a significant improvement of the efficiency of our algorithm. Figure 6 presents the average distance traveled by the search agent until the optimal path was found, as a function of the window size. I-A*DFS was employed for the low-level algorithm. The results shown in Figure 6 indicate that using a window of size larger than 1 (which corresponds to standard A*) significantly improves the algorithm's performance for the various graph sizes that we have experimented with. Also, we have found that the optimal size of the window tends to vary with the size of the graph. Based on our empirical observations, setting the optimal window size to (1/50) times the number of nodes in the graph seemed to provide a very good approximation. (For example, the best window sizes observed for 500- and 2000-node graphs were 10 and 40, respectively.) Note that as the window size becomes larger (i.e., the number of candidate nodes increases), the algorithm tends to select nodes with a large $f$ value, which results in performance degradation. Additional results for sparse and dense Delaunay graphs are presented in Appendix A.

At a first glance, the improvement of WinA* over standard A* (for the high level) seems somewhat modest, as it does not exceed 30%. This is due to the fact that I-A*DFS explores many nearby nodes, and is already very powerful to begin with. Both WinA* and I-A*DFS are designed to assign high priority to nearby nodes. They do so at different stages of the PHA* algorithm, but in a sense they "compete" for the same type of improvement.

Indeed, using any of the other navigating algorithms, the improvement of WinA* relative to standard A* was much more significant. However, in dealing with real physical agents — let alone humans — even the 30%-time reduction by WinA* (relative to I-A*DFS) should be viewed as significant. Similar results were obtained for sparse and dense Delaunay graphs (see Appendix A).

## 5. Analysis of PHA*

Analyzing the performance of PHA*, we distinguish between the following three parameters: (1) Cost of returned path, (2) shortest possible path that the agent can travel, and (3) cost of actual path traveled by the agent. In Subsection 5.1 we argue that the path reported by PHA* (for future use) is optimal. In addition, we present in Subsection 5.2 an extensive empirical study that compares between (2) and (3). Finally, we provide in Subsection 5.3 a brief discussion of PHA*'s underlying methodology and overall performance.

### 5.1 Optimality of Solution

Recall that A* expands nodes in a best-first order according to their $f$ value. If the heuristic function, $h(n)$, is admissible, then $f(n) = g(n) + h(n)$ is a lower bound on a path to the goal via node $n$. It is well-known, under this paradigm, that once the goal node is selected for expansion, A* has found an optimal path (Hart et al., 1968; Karp & Pearl, 1983; Dechter & Pearl, 1985). Put differently, if upon goal expansion $f(\text{goal}) = c$, then all other nodes with estimated paths of $f(n) < c$ have already been expanded and the length of the optimal path to the goal is $c$ (Karp & Pearl, 1983; Dechter & Pearl, 1985).

PHA* is supervised by the high level, which activates an admissible A*. (Recall that $h(n)$ is the Euclidean distance from $n$ to the goal, i.e., it is admissible.) By design of the algorithm, the high level terminates once the goal node is selected for expansion. Thus by the properties of admissible A*, all the nodes having a smaller $f$ value must have already been expanded, and the $f$ value of the goal is optimal. Note that this also holds for enhanced PHA* with WinA* (see Subsection 4.2.3). Although WinA* does not necessarily expand nodes according to the best $f$ value, it is designed to remove a node from the open list only if it has the smallest $f$ value among the nodes on the list. The algorithm halts only after the goal node has been expanded and removed from the open list, implying that its $f$ value is the smallest on the list. Thus our enhanced PHA* variant is also compatible with the admissible A* paradigm, and the path it returns is optimal. The basic theoretical result of the paper follows.

**Theorem:** *PHA* and its enhanced versions return the optimal path between the start node and the goal.*

### 5.2 Performance Evaluation of PHA*

As demonstrated above, the more complex algorithmic schemes provided a dramatic improvement in search time. It is of interest to assess, at least to some extent, the performance of our best navigation variant, i.e., WinA* (for the high level) in conjunction with I-A*DFS (for the low level).

| graph size | closed nodes | \|TSP$'$\| | PHA* | ratio |
|------|------|------|------|------|
| 30  | 11.32 | 0.62 | 0.80 | 1.29 |
| 50  | 15.45 | 0.74 | 0.94 | 1.27 |
| 75  | 17.93 | 0.77 | 0.97 | 1.22 |
| 100 | 20.32 | 0.85 | 1.10 | 1.29 |
| 150 | 24.12 | 0.91 | 1.27 | 1.39 |
| 200 | 28.43 | 0.99 | 1.42 | 1.43 |
| 250 | 31.57 | 1.02 | 1.48 | 1.45 |
| 300 | 35.78 | 1.05 | 1.51 | 1.44 |

Table 1: Comparison between shortest paths through nodes in closed list and actual paths obtained by PHA*.

The agent's task is to visit essentially all the nodes that are expanded by A*. These nodes comprise the set of nodes that are in the closed list when the algorithm terminates. In general, invoking A* on the subgraph induced by these nodes, with the same source and goal states and the same heuristic function, will exhibit the same behavior and yield the same open and closed lists. Thus given a static graph, the set of nodes which A* should visit is fixed. Ideally, we would like the agent to visit this set of *closed* nodes along the shortest possible path. This is of course infeasible, since the nodes are not known in advance, but rather determined on the fly. However, in order to evaluate our algorithm's performance, we can compare its output with the shortest possible path that travels through all these nodes. The computation of the latter is carried out off-line, i.e., after the set of (closed) nodes is known.

Specifically, we have computed the shortest possible path in each case with respect to the complete graph of the corresponding set of closed nodes. The weight $w(n_i, n_j)$ associated with an edge $(n_i, n_j)$ (in the complete graph) was set to the length of the shortest path from $n_i$ to $n_j$ (in the original Delaunay graph instance). Finding the shortest path that travels via a given set of nodes is known as the traveling salesman problem (TSP), which is notorious for its exponential running time. A conventional TSP path travels through all the nodes and then returns to the start node. However, we are interested in a path that travels through all the nodes without returning to the start node. We denote this path by TSP$'$ to distinguish it from the conventional TSP tour. A TSP$'$ tour is actually a TSP tour without the last edge. In view of the exponential nature of the problem, we have used a simple branch-and-bound tree search to compute the desired paths. However, solving this problem optimally was feasible only for relatively small graph sizes.

Table 1 provides a comparison between PHA* and the shortest path that travels through all the closed nodes for various small sized graphs. The table indicates that our PHA* algorithm is quite efficient for small graphs. Specifically, the average travel cost (due to PHA*) was not greater than the shortest possible path (passing through all the closed nodes) by more than 45%. For graphs having 200 nodes or less, the number of closed nodes observed was smaller than 30. The average cost in these cases was computed over 50 random instances. For graphs sizes greater than 200, the average cost was computed over 5 instances only.

In order to evaluate, however, the performance of **PHA\*** for graphs of larger size (where the optimal path could not be computed in a reasonable amount of time), we employed a lower-bound approximation to the cost of $TSP'$. Specifically, we have computed a minimum spanning tree (MST) of the complete graph (defined by the set of closed nodes). Let $|TSP'|$ and $|MST|$ denote, respectively, the costs associated with the desired path and the minimum spanning tree.

**Claim:**

$$0.5 \cdot |TSP'| < |MST| \leq |TSP'|.$$

**Proof:** The claim follows from basic graph theory (Cormen, Leiserson, Rivest, & Stein, 2001). Specifically, the inequality on the right hand side stems from the fact that $TSP'$ is a spanning tree of the complete graph. Thus the cost of a minimum spanning tree must be smaller than (or equal to) $|TSP'|$.

To prove the inequality on the left hand side, we note that the triangular inequality holds with respect to the above defined complete graph. (That is, for any three nodes, $n_i$, $n_j$, and $n_k$, $w(n_j, n_k) \leq w(n_i, n_j) + w(n_j, n_k)$.) This can be easily shown, based on the fact that the triangular inequality holds with respect to the original Delaunay graphs and by definition of an edge weight in the complete graph. Thus we can construct a tour that goes twice around the MST and then use the triangular inequality to shortcut some of its edges. Hence

$$2 \cdot |MST| \geq |TSP| > |TSP'|,$$

and the inequality on the left hand side follows. $\square$

Given the infeasible computation of $|TSP'|$, the claim suggests $|MST|$, instead, as a reasonably good approximation. Specifically, the inequality on the right hand side implies that if the travel cost of the agent performing PHA\* is, say, $c \cdot |MST|$, then the travel cost of PHA\* is no greater than $c \cdot |TSP'|$. Given that this is merely a lower bound, PHA\* is expected to perform better in practice.

Table 2 provides a comparison between PHA\* and the MST lower bound on the shortest path as described above. The average cost entered for each graph size was computed over 250 randomly generated instances. The table indicates that, on the average, the cost of PHA\* is at most 2.74 times that of the best possible path for graph sizes up to 8000 nodes and corresponding sets of closed nodes of up to 460 nodes.

## 5.3 Discussion

As was repeatedly noted, any algorithm that returns the optimal solution must expand at least all the nodes that are expanded by A\*. Drawing on this basic premise, our PHA\* algorithm was designed to visit the set of "mandatory" nodes as efficiently as possible. The rationale of visiting also nearby nodes (whose $f$ value is not necessarily the smallest) is that such nodes are likely to be expanded in the next few iterations. In contrast, there is no benefit to this enhanced variation in the context of a navigation algorithm that does not presume to return an optimal solution.

| graph size | closed nodes | \|MST\| approx. | PHA* | ratio |
|---|---|---|---|---|
| 400 | 40.27 | 1.05 | 1.91 | 1.82 |
| 500 | 43.00 | 1.15 | 1.97 | 1.87 |
| 1000 | 62.72 | 1.42 | 3.03 | 2.13 |
| 2000 | 131.56 | 2.01 | 4.89 | 2.43 |
| 4000 | 233.26 | 2.52 | 6.76 | 2.69 |
| 8000 | 460.66 | 3.45 | 9.44 | 2.74 |

Table 2: Comparison between lower bounds on shortest paths through nodes in closed list and actual paths obtained by PHA*.

Reconsider Roadmap-A*, for example. $A_\varepsilon^*$ is only activated to prevent the local navigation phase from going in the wrong direction. However, since this algorithm is not designed to return an optimal solution, it will not deviate at any stage from its promising route to visit a nearby node that may be expanded later on. Put differently, there is no notion here of a set of mandatory nodes that the agent has to visit. Furthermore, as soon as the agent reaches the goal, the search halts. In conclusion, although both PHA* and Roadmap-A* are two-level navigation schemes, their objectives are different and they solve essentially different problems.

Based on the properties of admissible A* and by design of our algorithm, we have argued that (enhanced) PHA* returns a path (for future use) that is optimal. In addition, in the absence of a theoretically known bound on the actual cost of PHA*, we have run an extensive empirical study, comparing between observed costs and the best possible costs computed off-line. Given that the agent lacks a priori information as to the set of mandatory nodes, it is highly unlikely that there exists an on-line PHA*-like algorithm that performs as efficiently as an off-line version. Our extensive empirical study demonstrates, nevertheless, that the actual cost associated with PHA* is on the same order of magnitude as the optimal cost computed off-line.

## 6. MAPHA*: Multi-Agent PHA*

In this section we generalize the techniques discussed in the previous sections to the multi-agent case, where a number of agents cooperate in order to find the shortest path. We call the resulting algorithm Multi-Agent Physical A* (MAPHA*).

We would like to divide the traveling effort between the agents in the most efficient way possible. We can measure this efficiency for the multi-agent case using two different criteria. The first is the overall global time needed to solve the problem. The second is the total amount of fuel that is consumed by all agents during the search. If the requirement is to minimize the cost of moving the agents and time is not important, then considering the fuel cost of mobilizing the agents will be the cost function of choice. In this case, it may be wise to move some agents while other agents remain idle. However, if the task is to find the best path to the goal, as soon as possible, idle agents seem wasteful, as they can better utilize their time by further exploration of the graph. In such a case, all available agents should be moving at all times. We introduce below two algorithms for these two perspectives, namely a *fuel-efficient* algorithm and a *time-efficient* algorithm. Note that in the single agent case these two criteria coincide.

We assume that each agent can communicate freely with all the other agents and share data at any time. Thus any information gathered by one agent is available and known to all of the other agents. This framework can be obtained by using a model of a centralized supervisor that moves the agents according to the complete knowledge that was gathered by all of them. This is a reasonable assumption since in many cases there is a dispatcher or some centralized controller that gathers information from the agents and instructs them accordingly. Another possible model for complete knowledge-sharing is that each agent broadcasts any new data about the graph to all the other agents. Future research may address a more restrictive communication model, by limiting the communication range or inducing communication errors.

We also assume that the search terminates, as soon as the goal node is expanded and moved to the closed list. Our objective is to minimize the travel effort up to that point, and we do not care about moving all the agents to some pre-specified location (e.g., the goal vertex or the start node), after the desired shortest path is identified. This convention is in accordance with many algorithms which neglect to report the time spent to "reset" a system (e.g., garbage collection), once the desired solution is arrived at.

The main idea of the MAPHA* algorithm is very similar to that of PHA* for a single agent. We use again a two-level framework. The high level chooses which nodes to expand, while the low level navigates the agents to these nodes. We have studied the multi-agent case with our enhanced techniques only, i.e., WinA* for the high level and I-A*DFS for the low level. The problem that we deal with below is how to assign the different agents to explore efficiently the different nodes.

## 6.1 MAPHA*: Fuel-Efficient Algorithm

For simplicity, we assume that the amount of fuel consumed by an agent is equal to its traveling distance during the search. Since the purpose of the algorithm in this case is to minimize the amount of fuel consumed by the agents, regardless of the overall search time, there is no benefit to moving more than one agent at a time. This is because by moving only one agent, that agent might gain new knowledge of the graph that would allow the other agents to make more informed and intelligent moves.

At the beginning, all the agents are situated at the source node. Then, as in the case of a single agent, the high level defines a window of unexplored nodes from the open list that are potential candidates for expansion. For each pair $(a, n)$, where $a$ is an agent and $n$ is a node from the window, we compute the allocation cost function

$$c(a, n) = f(n) \cdot \text{dist}(a, n),$$

where $f(n)$ is the $f$ value of node $n$ and $\text{dist}(a, n)$ denotes the distance from the location of agent $a$ to node $n$. We then select an agent and a target node that minimize that allocation function. In the case of tie-breaking (e.g., at the beginning of the search where all agents are located at the initial state), we pick randomly one agent from the relevant candidates. At this stage, the low-level algorithm navigates the selected agent to the target node selected from the window in order to explore that node. As in the single-agent case, additional knowledge about the graph is being obtained during the navigation as many unexplored nodes are visited by the traveling agent. Only when the selected agent reaches its target is a new cycle activated for the high- and low-level procedures.[9] Following is the pseudo-code for the fuel efficient algorithm.

```
fuel-efficient algorithm() {
.    while (goal is not in closed-list) {
.        for each agent a_i
.            select node n_i from the window that minimizes f(n) · dist(a_i, n);
.        a_best = agent that minimizes f(n_i) · dist(a_i, n_i);
.        if n_best is unexplored then
.            explore(n_best) by low level using a_best;
.        expand(n_best);
.        while (best node in open-list was expanded)
.            close(best node);
.    }
}
```

## 6.2 MAPHA*: Time-Efficient Algorithm

The time-efficient algorithm is very similar to the above described fuel-efficient algorithm with one basic modification. Instead of moving only one agent during each high-level cycle, we now move all of the available agents since we only care about the time spent by the agents and not about their fuel consumption. Having an idle agent will not save any time. Every moving agent can only help gather more knowledge about the environment with no additional cost, as the clock ticks away regardless and the time is measured globally.

We cannot use here the same allocation function that was used for the fuel-efficient algorithm, as all agents are located initially at the same node, and the fuel-efficient allocation function will choose the same node for all the agents. The main idea of the time-efficient strategy is that all agents move simultaneously. Thus to ensure efficient performance we need to distribute them as much as possible. Suppose that we have $p$ available agents and $k$ nodes in the window. We would like to distribute these $p$ agents to the $k$ nodes as efficiently as possible. A brute-force approach will be to randomly distribute the agents to the nodes. However, to provide an effective distribution, we incorporate the following three criteria into our distribution formula for the time-efficient procedure:

1. Since the $f$ values of neighboring nodes are somewhat correlated with each other, nodes with a small $f$ value are more likely to generate new nodes with a small $f$ values than nodes with a large $f$ value. Therefore, the distribution should favor assigning an agent to a node with a small $f$ value.

2. Another attribute that should be taken into consideration is the distance of the target node from an agent. We would like to assign an agent to one of the nodes in such a manner, that the expected travel distance of the agent (for that assignment) is minimized. In other words, an agent will be assigned, preferably, to a relatively close-by node.

3. In order to expand the entire window and prevent "starvation", we would also like our distribution function to raise the priority of nodes that were assigned a small number of agents. Thus we should keep track of the number of agents that were assigned to each node and give preference to nodes with a small number of assignments.

Note that the first and third criteria may contradict, i.e, the first criterion will prefer nodes with a small $f$ value while the third criterion will favor nodes with a large $f$ value, as only a small number of agents was assigned to them.

We have found that taking the product of the values associated with these three criteria gives a good distribution function with a suitable load balancing between these criteria. Specifically, the agent allocation procedure iterates through all the agents and picks, for each agent, the node that minimizes the following allocation function:

$$\text{alloc}(\text{agent}, \text{node}) = f(\text{node}) \cdot \text{dist}(\text{agent}, \text{node}) \cdot (\text{count}(\text{node}) + 1),$$

where $\text{dist}(\text{node}, \text{agent})$ is the Euclidean distance between the node and the agent, $f(\text{node})$ is that node's $f$ value, and $\text{count}(\text{node})$ is a counter that keeps track of the number of agents that have already been assigned to explore that node. $\text{count}(\text{node})$ is initially set to 0 and is incremented every time an agent is assigned to that node. Thus a load balancing between the three factors is being kept throughout the distribution process. At the beginning of the search all the agents are located at the start node, and their initial allocation to the different nodes is determined, essentially, by the count factor. (Without this factor, the product $f(n) \cdot \text{dist}(\text{agent}, n)$ would have returned the same node $n$ for all agents.) As the search progresses, the agents move to different locations and get assigned at each step to nodes that are closer to their location and that have a small $f$ value. Thus the product of the above three factors creates a good distribution (of the agents) over different parts of the graph.

Consider, for example, the case illustrated in Figure 7. Suppose that 100 agents are all located at node $x$, and that the window consists of the three nodes $a$, $b$, and $c$ that are located at an equal distance from $x$. Suppose also that $f(a) = 2$, $f(b) = 4$ and, $f(c) = 8$. The numbers of agents that are assigned to these nodes, using the above allocation procedure, are 57, 29, and 17, respectively. This is a good balance between the various requirements.

We have tried many other variations for the distribution procedure and found that they all performed well as long as the above three requirements were met. See (Stern, 2001) for further discussion on agent distribution.

As before, each agent navigates to its assigned target using our enhanced low-level algorithm, I-A*DFS. Another high-level iteration begins as soon the the first agent reaches its target node.[10] Note that the computation time of the window and that of the agent distribution/allocation can be neglected, since we only care about the travel time of the agents. Following is the pseudo code for the time-efficient algorithm.

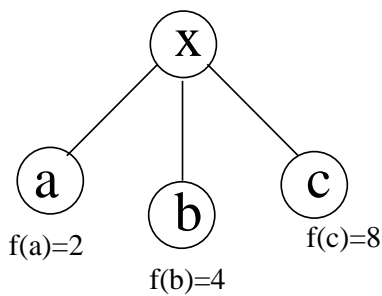

Figure 7: An example of agent distribution according to the proposed allocation procedure.

```
time-efficient algorithm() {
.    while (goal is not in closed-list) {
.        for each free agent a_i
.            select a window node n_i that minimizes dist(a_i,n)·f(n)·(count(n+1));
.        move all agents until an agent reaches a node;
.        expand all nodes currently visited by an agent;
.        while (best node in open-list was expanded)
.            close(best node);
.    }
}
```

## 6.3 Experimental Results

The experiments performed for the multi-agent case were also conducted on Delaunay graphs with 500, 1000, 2000, 4000, and 8000 nodes. Additional results for sparse and dense Delaunay graphs are provided in Appendix A.

### 6.3.1 MAPHA*: Results for the Fuel-Efficient Algorithm

We provide here results for the fuel-efficient algorithm of Subsection 6.1. The fuel consumption reported is the total fuel consumed by all the agents. (As before, the graphs were generated on a unit square, for which the average optimal path observed was about 0.55.)

Figure 8 presents the costs of the fuel-efficient algorithm as a function of the number agents for various sizes of regular Delaunay graphs. (Results for sparse graphs, as well as graphs with edges added at random, are presented in Appendix A.) The figure clearly demonstrates that as more agents are added, the overall fuel consumption decreases up to a point where adding more agents tends to increase the overall consumption. Thus an optimal number of agents exists for each of the graphs. This phenomenon is due to the fact that A* is usually characterized by a small number of search regions. Therefore, a small number of agents suffices to cover these regions. As the number of agents increases, the fuel consumption goes up. This phenomenon is explained as follows. A large number of agents increases the likelihood that a nearby agent will be assigned to a specific node, in which case relatively little exploration of the graph takes place. Assigning, on the other hand, a distant agent to the node would result in a larger degree of graph exploration, which is essential, in the long run, for efficient navigation (especially if I-A*DFS is employed). Thus a large number of agents navigating in a small graph (which has few search regions), would result in excessive fuel consumption. See (Stern, 2001) for a more detailed explanation of this phenomenon.

---

10. We have observed that when a new iteration begins, almost every agent is assigned to the same node that it was assigned to in the previous iteration. Typically this is because an agent's location becomes closer to "its" target node, while the other criteria do not change. Thus in practice, most of the agents go on to complete their (original) tasks, and only the agent that has reached its target is assigned a new goal node. See (Stern, 2001) for a detailed discussion.

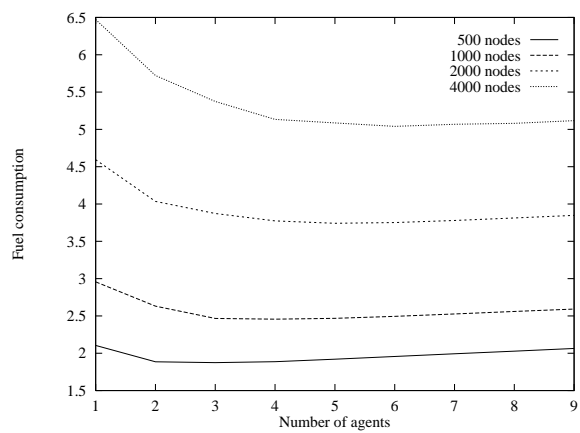

Figure 8: Fuel consumption as a function of number of agents for various sizes of regular Delaunay graphs.

The optimal number of agents increases as the number of nodes in the graph increases. While the optimal number of agents for a graph of 500 nodes is 2, this number increases up to 7 for a graph of size 4000. This stems from the fact that larger graphs have more search regions and thus more agents are needed to explore them.

As described before, only one agent is allowed to move in this experiment, at any point in time. Up to now we have measured the total amount of fuel consumed by all of the agents. It is of interest to find out whether the work is uniformly distributed among the agents, or whether a large portion of the work is carried out by a small number of agents. Table 3 presents the distribution of the work among the agents when up to 14 agents were active on Delaunay graphs of size 8000. For each graph instance, we sorted the agents in decreasing order of their fuel consumption. The table shows the relative fuel consumption of the agents for 3, 7, and 14 activated agents.

In general, we remark that while the overall work is not uniformly distributed, it is quite balanced. For example, when 14 agents are activated, 40% of the work is done by only 4 agents. A similar tendency was observed for graphs of other sizes, as well as for sparse and dense Delaunay graphs (see Appendix A).

| Agent No. | 3 agents [%] | 7 agents [%] | 14 agents [%] |
|:---:|:---:|:---:|:---:|
| 1 | 42.03 | 28.34 | 16.01 |
| 2 | 33.54 | 19.32 | 13.21 |
| 3 | 24.43 | 16.23 | 11.41 |
| 4 | | 12.76 | 10.31 |
| 5 | | 9.02 | 6.64 |
| 6 | | 7.86 | 5.48 |
| 7 | | 6.48 | 5.08 |
| 8 | | | 4.54 |
| 9 | | | 4.10 |
| 10 | | | 3.59 |
| 11 | | | 3.20 |
| 12 | | | 3.02 |
| 13 | | | 2.81 |
| 14 | | | 2.70 |

Table 3: Work distribution among multiple agents running our fuel-efficient algorithm on Delaunay graphs of size 8000.

In order to improve the efficiency of the fuel-efficient algorithm and to make the overall work distribution more balanced, several improvements might be suggested. For example, currently all the agents are positioned initially at the same source node. We might consider to first spread the agents in a number of directions and only then invoke the algorithm. Notwithstanding the additional overhead that may be incurred by spreading the agents, this technique can result in a more balanced work distribution and in a reduced overall fuel consumption.

### 6.3.2 MAPHA*: Results for the Time-Efficient Algorithm

In this subsection we report the results for the time-efficient algorithm of Subsection 6.2. As was explained, if our main objective is to conclude the task as fast as possible, such that fuel consumption is of no concern, then all agents should always be moving, i.e., none of them should be idle at any point in time. The overall search time in this case is the maximal distance that either agent travels until the shortest path to the goal node is found.

Figure 9 shows the search time obtained by the time-efficient algorithm as a function of the number of agents, for various regular Delaunay graphs. Note that the search time can never be smaller than the time it takes to travel along the shortest path to the goal. As the results indicate, adding more agents is always efficient since we only measure the overall time that has elapsed until the goal is found. What makes our algorithm interesting and efficient is the fact that as we add more agents, the search time converges asymptotically to the length of the shortest path. Recall that the average length observed of the shortest path was approximately 0.55. Indeed, a large number of agents will tend to find the optimal path within a time frame that approaches the above limit. While the overall time was 2.3 with a single agent, it was reduced to 0.7 with 14 agents for graphs with 500 nodes for example.

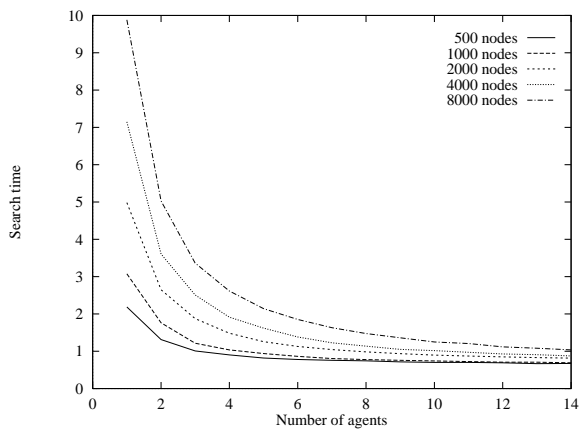

Figure 9: Time consumption as a function of the number of agents, for various regular Delaunay graphs.

Using our proposed agent allocation procedure, we note that asymptotically all paths from the initial state are traveled in a breadth-first search manner. This is to say that a sufficiently large team of agents is likely to produce a single agent that will travel along the actual shortest path with very little deviation from it. Similar results for the time- efficient algorithm were also obtained for other types of graphs (see Appendix A).

## 6.4 Combined Requirements of Fuel and Time

While the distinction between the time-efficient algorithm and fuel-efficient algorithm is reasonable, it may not be suitable in many practical situations. Practical considerations of time and fuel resources may suggest a combined approach, as the one described below.

Consider, for example, a commander operating under a constraint of fuel consumption but with no restriction on the number of troops that can be assigned to a certain task. In order to complete the task as fast as possible, the commander may want to use the maximal possible number of agents without exceeding the fuel consumption limit.

In essence, we seek to generalize MAPHA*, such that the agents will minimize a cost function which is a combination of time and fuel consumption. We suggest a general cost function that takes into account the requirements on both these measures. The objective will be to activate MAPHA*, so as to minimize this cost function. Specifically, we suggest the following linear combination:

$$C_{\text{total}} = w_t \cdot \text{time} + w_f \cdot \text{fuel},$$

where $w_t$ and $w_f$ are the (normalized) weights attached, respectively, to the time and fuel consumption (i.e., $0.0 \leq w_t, w_f \leq 1.0$ and $w_t + w_f = 1.0$). $C_{\text{total}}$ is calculated globally, i.e., we measure the amount of time from the beginning of the task until the optimal path is found, and the total amount of fuel consumed by all the agents. We then multiply these quantities by their corresponding weights and report the combined cost.

Both $w_t$ and $w_f$ are prespecified by the user. If $w_t = 0$, there is no time cost and the fuel-efficient algorithm will be the appropriate one to use. If $w_f = 0$, there is no fuel cost, and we can use the time-efficient algorithm. Otherwise, if neither $w_t$ nor $w_f$ is 0, we should use a different algorithm to minimize $C_{\text{total}}$.

We suggest two algorithms for this general case.

- Simple combined algorithm.

  This algorithm is actually identical to the time-efficient algorithm. The number of participating agents is a parameter provided by the user. At each iteration of the high level all the participating agents move according to the allocation function of the time-efficient algorithm. Given the formulation of the total cost, $C_{\text{total}}$, we would like to determine the optimal number of agents, for any $w_t$ and $w_f$. Note that in the trivial case where $w_f = 0$, adding more agents is always valuable, since they do not consume any resources, and can only reduce the time cost. However, as $w_f$ increases, a large number of agents may increase the total cost.

- Improved combined algorithm.

  The main limitation of the simple combined algorithm is that even though cost is incurred for fuel consumption, all the agents are always moving. The *improved combined algorithm* addresses this problem and suggests moving only some of the agents simultaneously. Using this formalization, we first determine $p$, i.e., the number of agents that will participate in the task. Given $p$, we then determine $m$, i.e., the number of agents that will actually be distributed to nodes selected from the window (by the high level). The remaining $p - m$ agents will stay idle. Note that for the simple combined algorithm $p$ an $m$ coincide. We use the same mechanism of the time-efficient allocation function, except that here the algorithm chooses only $m$ (out of $p$) agents that minimize this allocation function. As in the time-efficient algorithm, we first determine the size of the window, i.e., the number of nodes from the open list that should be expanded. Then, we invoke the same allocation function. Whereas in the time-efficient case the allocation terminates once all the agents are assigned to nodes, here the allocation stops after $m$ agents are selected. The selected agents are the best $m$ agents for this expansion cycle since they minimize the allocation function.

### 6.5 Results for the Combined Algorithm

We provide experimental results for the combined algorithm that was introduced in the previous subsection. The results in Tables 4 and 5 were obtained for Delaunay graphs of size 2000; each table entry represents the average of 250 problem instances. For each column, the bold face number is the smallest total cost for the corresponding $w_t/w_f$ ratio. These minimal costs determine the optimal number of agents for a given $w_t/w_f$ ratio.

Table 4 provides total costs for the simple combined algorithm as a function of the number of agents for various $w_t/w_f$ ratios. The leftmost column corresponds to the case where only time matters. Thus its entries are identical to the values obtained by the time-efficient algorithm. As fuel consumption becomes more significant, it is no longer beneficial to increase the number of agents and thus the optimal number of agents decreases. For $w_t = w_f = 0.5$, $C_{\text{total}} = 0.5 \cdot$ time + $0.5 \cdot$ fuel, the optimal number of agents obtained is three, for a total cost of 3.49. The more critical fuel consumption becomes, the more beneficial it is to use a smaller number of agents. The rightmost column corresponds to the other extreme case, where $w_f = 1.0$, i.e., when only fuel consumption matters. Note that the entries of this column differ from their counterpart costs obtained by the fuel-efficient algorithm. The difference stems from the fact that, in the context of the simple combined algorithm, picking $p$ agents means that they will all be moving simultaneously, whereas in case the fuel-efficient algorithm is employed only one agent (out of $p$) will be allowed to move at all times. Note that the fuel-efficient algorithm is essentially a special case of the improved combined algorithm with $m = 1$.

| $w_t$ | 1.0 | 0.9 | 0.8 | 0.7 | 0.6 | 0.5 | 0.4 | 0.3 | 0.2 | 0.1 | 0.0 |
|---|---|---|---|---|---|---|---|---|---|---|---|
| $w_f$ | 0.0 | 0.1 | 0.2 | 0.3 | 0.4 | 0.5 | 0.6 | 0.7 | 0.8 | 0.9 | 1.0 |
| # agents | | | | | | | | | | | |
| 1 | 4.82 | 4.82 | 4.82 | 4.82 | 4.82 | 4.82 | 4.82 | 4.82 | 4.82 | **4.82** | **4.82** |
| 2 | 2.58 | 2.84 | 3.10 | 3.36 | 3.61 | 3.87 | 4.13 | 4.39 | 4.65 | 4.91 | 5.16 |
| 3 | 1.74 | 2.09 | 2.44 | 2.79 | **3.14** | **3.49** | **3.84** | **4.18** | **4.53** | 4.88 | 5.23 |
| 4 | 1.44 | 1.87 | 2.30 | **2.73** | 3.16 | 3.60 | 4.03 | 4.46 | 4.89 | 5.32 | 5.75 |
| 5 | 1.24 | 1.73 | 2.23 | 2.72 | 3.22 | 3.71 | 4.21 | 4.70 | 5.20 | 5.69 | 6.19 |
| 6 | 1.11 | 1.67 | **2.22** | 2.78 | 3.33 | 3.89 | 4.44 | 5.00 | 5.55 | 6.11 | 6.66 |
| 7 | 1.03 | **1.64** | 2.26 | 2.88 | 3.49 | 4.11 | 4.73 | 5.34 | 5.96 | 6.58 | 7.19 |
| 8 | 0.97 | 1.65 | 2.33 | 3.01 | 3.69 | 4.37 | 5.05 | 5.73 | 6.41 | 7.09 | 7.77 |
| 9 | 0.93 | 1.68 | 2.42 | 3.17 | 3.91 | 4.66 | 5.40 | 6.15 | 6.89 | 7.64 | 8.38 |
| 10 | 0.89 | 1.70 | 2.50 | 3.31 | 4.11 | 4.92 | 5.72 | 6.53 | 7.33 | 8.14 | 8.95 |
| 11 | 0.85 | 1.70 | 2.56 | 3.41 | 4.26 | 5.11 | 5.96 | 6.81 | 7.67 | 8.52 | 9.37 |
| 12 | 0.84 | 1.77 | 2.70 | 3.63 | 4.56 | 5.49 | 6.42 | 7.35 | 8.27 | 9.20 | 10.13 |
| 13 | 0.84 | 1.84 | 2.84 | 3.84 | 4.84 | 5.85 | 6.85 | 7.85 | 8.85 | 9.86 | 10.86 |
| 14 | **0.82** | 1.88 | 2.94 | 4.01 | 5.07 | 6.13 | 7.19 | 8.26 | 9.32 | 10.38 | 11.44 |

Table 4: $C_{\text{total}}$ for the simple combined algorithm as a function of the number of agents, for various ratio $w_t/w_f$ ratios.

Table 5 provides total costs for the improved combined algorithm as a function of the number of agents for various $w_t/w_f$ ratios. The number of participating agents was $p = 14$ (i.e., up to 14 available agents could move simultaneously). Each row corresponds to a different $m$, i.e., to the actual number of moving agents. (Clearly, $1 \le m \le p = 14$.) As before, for each column the bold face number is the smallest total cost for the corresponding $w_t/w_f$ ratio. These minimal costs determine the optimal number of moving agents for a given $w_t/w_f$ ratio.

The top entry of the rightmost column is identical to the cost obtained by the fuel-efficient algorithm, for 14 agents. In this case $w_f = 1$, and only one agent is allowed to move at any point in time. The bottom entry of the leftmost column is identical to the cost obtained by the time-efficient algorithm, for 14 agents. In this case $w_t = 1$, and all of the 14 participating agents are moving at all times.

The more significant fuel consumption becomes, the less beneficial it is to move many agents. Thus the optimal number of moving agents decreases. For example, for $w_t = w_f = 0.5$, the optimal number of moving agents obtained was three, for a total cost of 3.23. As fuel consumption becomes more crucial, it would be beneficial to move a smaller number of participating agents.

Comparing the results of the simple combined algorithm with those of the improved combined algorithm reveals that for the same $w_t/w_f$ ratio and for the same number of moving agents (which is equal to the number of all participating agents in the simple combined case) the improved combined algorithm usually performs better. This is because it can pick the moving agents from a larger sample. Also, it appears that the optimal number of moving agents is smaller for the improved combined algorithm. In this algorithm, the moving agents are picked in a clever manner at each cycle and thus can be better utilized.

Additional experiments were conducted for other graph sizes, as well as for sparse and dense Delaunay graphs. The results obtained in all cases were rather consistent. Future work will attempt to predict in advance the best number of agents.

## 7. Conclusions and Future Work

We have addressed the problem of finding the shortest path to a goal node in unknown graphs that represent physical environments. We have presented the two-level algorithm, PHA*, for such environments for a single search agent, and the MAPHA* algorithm for multiple-agents. We have experimented with several variations of Delaunay graphs, containing up to 8000 nodes. The enhanced single agent algorithm yielded significantly better results than the ones obtained by the simpler variants. The results for the fuel-efficient algorithm show that using more agents is beneficial only to some extent. This is because all the agents are initially located at the same source node and they all consume fuel for each move they make. For the same reason, the benefit of using the optimal number of agents as opposed to only one agent is modest. The results for the time-efficient algorithm are very encouraging, since the search time converges quickly to the optimum as the number of search agents increases. We have also introduced a cost function that combines both time consumption and fuel consumption, and have presented two algorithms for this paradigm. The results show that for each combination there exists an optimal number of agents which tends to increase as the weight of the time cost increases.

| $w_t$ | 1.0 | 0.9 | 0.8 | 0.7 | 0.6 | 0.5 | 0.4 | 0.3 | 0.2 | 0.1 | 0.0 |
|---|---|---|---|---|---|---|---|---|---|---|---|
| $w_f$ | 0.0 | 0.1 | 0.2 | 0.3 | 0.4 | 0.5 | 0.6 | 0.7 | 0.8 | 0.9 | 1.0 |
| # agents | | | | | | | | | | | |
| 1 | 4.02 | 4.02 | 4.02 | 4.02 | 4.02 | 4.02 | 4.02 | 4.02 | **4.02** | **4.02** | **4.02** |
| 2 | 2.26 | 2.49 | 2.72 | 2.94 | 3.17 | 3.40 | 3.62 | **3.85** | 4.08 | 4.30 | 4.53 |
| 3 | 1.61 | 1.94 | 2.26 | 2.58 | 2.91 | **3.23** | **3.55** | 3.88 | 4.20 | 4.52 | 4.84 |
| 4 | 1.31 | 1.70 | 2.09 | 2.49 | **2.88** | 3.27 | 3.67 | 4.06 | 4.45 | 4.84 | 5.24 |
| 5 | 1.13 | 1.58 | 2.03 | **2.48** | 2.93 | 3.38 | 3.83 | 4.28 | 4.73 | 5.18 | 5.63 |
| 6 | 1.00 | 1.49 | **1.99** | 2.49 | 2.99 | 3.49 | 3.98 | 4.48 | 4.98 | 5.48 | 5.98 |
| 7 | 0.92 | 1.47 | 2.02 | 2.57 | 3.13 | 3.68 | 4.23 | 4.78 | 5.33 | 5.88 | 6.43 |
| 8 | 0.85 | **1.45** | 2.05 | 2.65 | 3.25 | 3.85 | 4.44 | 5.04 | 5.64 | 6.24 | 6.84 |
| 9 | 0.82 | 1.47 | 2.12 | 2.77 | 3.43 | 4.08 | 4.73 | 5.39 | 6.04 | 6.69 | 7.34 |
| 10 | 0.79 | 1.50 | 2.21 | 2.92 | 3.63 | 4.34 | 5.05 | 5.76 | 6.47 | 7.18 | 7.89 |
| 11 | 0.77 | 1.54 | 2.31 | 3.07 | 3.84 | 4.61 | 5.38 | 6.15 | 6.92 | 7.69 | 8.46 |
| 12 | 0.76 | 1.59 | 2.42 | 3.25 | 4.09 | 4.92 | 5.75 | 6.58 | 7.41 | 8.25 | 9.08 |
| 13 | 0.75 | 1.64 | 2.54 | 3.44 | 4.33 | 5.23 | 6.13 | 7.02 | 7.92 | 8.82 | 9.71 |
| 14 | **0.75** | 1.72 | 2.69 | 3.67 | 4.64 | 5.61 | 6.59 | 7.56 | 8.53 | 9.51 | 10.48 |

Table 5: Total costs for the improved combined algorithm as a function of the number of moving agents (out of 14 participating agents), for various $w_t/w_f$ values.

Future work can be pursued along the following directions:

- We have assumed that upon reaching a node, the agent can learn the locations of all of its neighbors. In many domains this model may not be valid, and the location of a node is known only when the agent actually visits it. Such a model was also suggested in (Shmoulian & Rimon, 1998). Further research should be done in order to implement our algorithms, in the context of such a model.

- We have used traveling agents to solve the shortest path problem. A similar mechanism might be used for solving other known graph problems, such as the minimum spanning tree, the traveling salesman problem, or any other problem that requires consideration as to which node should be visited next.

- We have proposed two algorithms for combining time consumption and fuel consumption. Both algorithms assume that the number of agents is determined a priori. Future work can try to theoretically determine the optimal number of agents given the constraints. Also, future work can be done to see whether changing this number on the fly would increase the efficiency of these algorithm. Also, we have assumed that agents consume fuel only when they move, and have measured only the total performance time of a task. Thus idle agents do not consume any resources. However, we can think of a model where idle agents do consume resources (e.g., time and energy).

- We have assumed a centralized model, where all the agents share their knowledge at all times. Future work can assume other communication paradigms. In particular, we are interested in a model where there is no communication at all between the agents. This model is known as the ant-robotics model (Wagner & Bruckstein, 2000; Yanovski, Wagner, & Bruckstein, 2001). In this model, information is spread to other agents by *pheromones*, i.e., data that are written by an agent at a node. Other agents can read these pheromones when reaching those nodes. We are currently working towards applying our MAPHA* algorithm to such a model. We believe that if we increase the size of the data that are allowed to be written at each node, then each agent will be able to write its complete knowledge at a node of the environment. The challenge of applying A* in such a model lies in the fact that since A* maintains a global open list, data from opposite sides of the graph can influence the behavior of the algorithm. Thus we need the knowledge sharing of such a system to be as large as possible. For this purpose, we believe that a new type of communication agents should be introduced. Agents of this type will not try to increase the search frontier but rather move around the environment and spread the most recent data available.

## Acknowledgments

A preliminary version of this paper appeared in *Proceedings of the First International Joint Conference on Autonomous Agents and Multi-Agent Systems*, 2002 (Felner et al., 2002). The work was carried out while the first author was at Bar-Ilan University. This material is based upon work supported in part by NSF under grant #0222914 and by ISF under grant #8008.

## Appendix A. Additional Experimental Results

As mentioned in Subsection 4.3, each node in a *regular* Delaunay graph is connected to all its neighbors. This property may not always apply to a real road map. For example, nearby geographic locations may not always be connected by a road segment, due to the the existence of obstacles like a mountain or a river. In addition, distant locations are often connected by highways. To capture these additional characteristics, we have also considered so-called *sparse* and *dense* Delaunay graphs. Instances of these variants can be easily obtained from regular Delaunay graphs by random deletion and addition of edges, respectively. Specifically, we have generated sparse Delaunay graph instances by deleting roughly 60% of the edges at random. Likewise, dense instances were generated by introducing 400 edges at random. (A new edge is created by selecting at random a pair of nodes.)

We have run all of the algorithms presented in the main body of the paper also on the above Delaunay graph variants. The results obtained are presented here.

As can be expected, the more sparse the graph, the more often the agent runs into deadends. Indeed, all the algorithms required additional travel effort to find the optimal path after edges were removed. However, the ratio between the travel cost of any two algorithms seems to remain the same (for the various Delaunay graph types), and I-A*DFS exhibits superior performance for all graph instances. See Figures 10(a), (b). This behavior proved consistent in all of our experiments, for both a single agent and a multi-agent environment. Also, Figures 11(a), (b) exhibit similar behavior of search cost of WinA* versus window size for sparse Delaunay graphs and dense Delaunay graphs, respectively, to that observed for regular Delaunay graphs (see Figure 6).

Figures 12(a), (b) present the costs of the fuel-efficient algorithm as a function of the number agents for various sizes of sparse and dense Delaunay graphs, respectively. The overall fuel consumption recorded for the sparse Delaunay graphs is larger than the fuel consumption recorded for their counterpart regular graphs (see Figure 8) by a factor of about 1.5. For graphs simulating highways (i.e., the dense graphs) the fuel consumption decreases relative to both sparse and regular Delaunay graphs.

Note that the optimal number of agents navigating in a sparse graph also increases, since agents need to backtrack more often in this case. Thus having more agents will assist the search. On the other hand, adding random edges to the graphs causes the opposite effect, i.e., less fuel is consumed and the optimal number of agents is reduced. This is explained by the fact that new edges add more connections between nodes, i.e., many "shortcuts" are created and the search can be carried out faster and with a smaller number of agents.

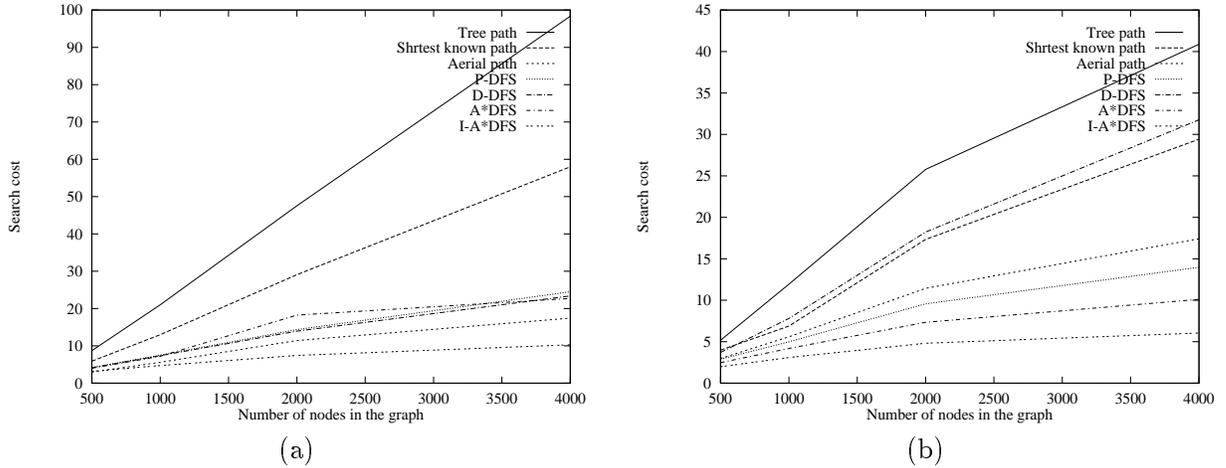

Figure 10: Search cost versus the number of nodes of: (a) Sparse Delaunay graphs, and (b) dense Delaunay graphs for various low-level algorithms.

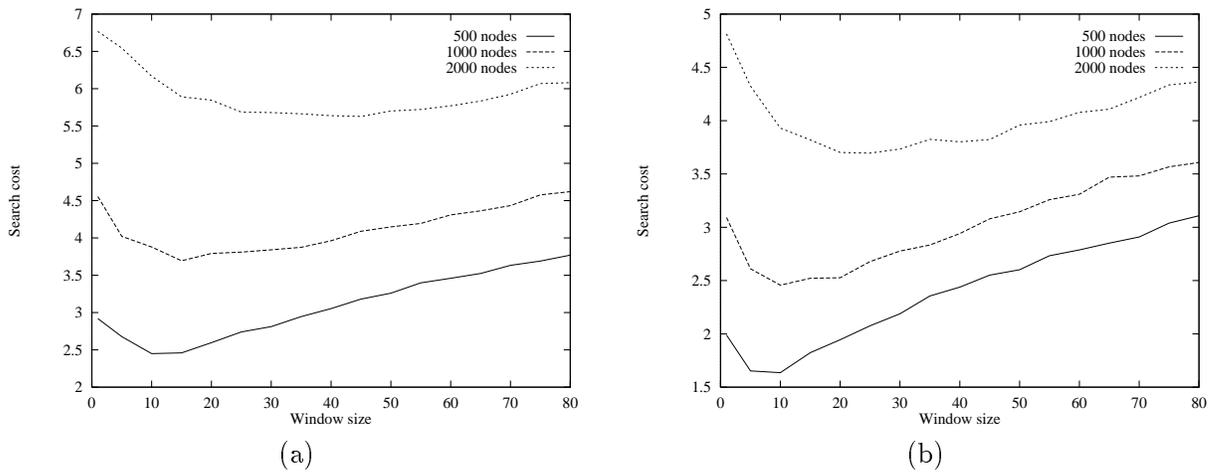

Figure 11: Search cost of WinA* versus window size for various sizes of: (a) Sparse Delaunay graphs, and (b) dense Delaunay graphs.

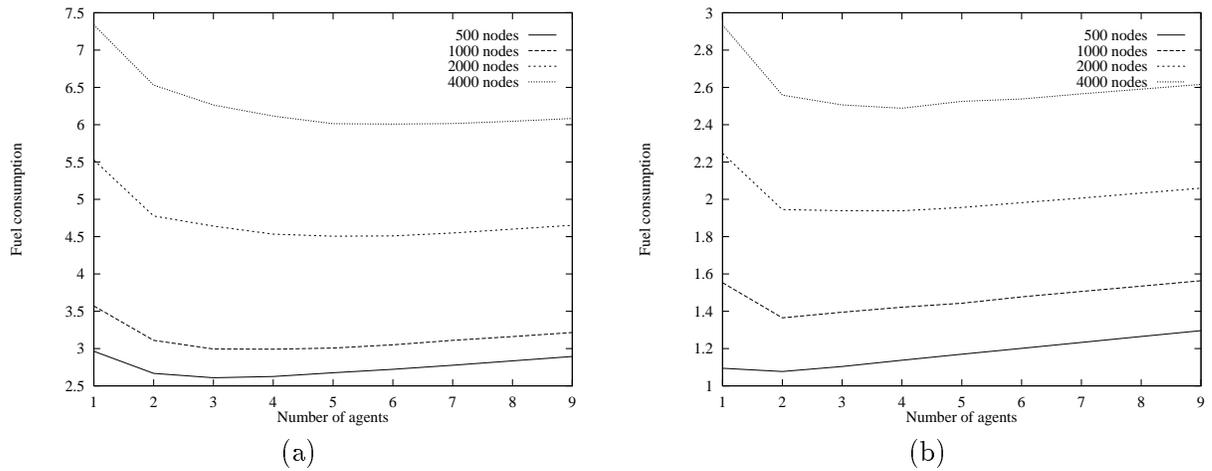

Figure 12: Fuel consumption as a function of the number of agents for various sizes of: (a) Sparse Delaunay graphs, and (b) dense Delaunay graphs.

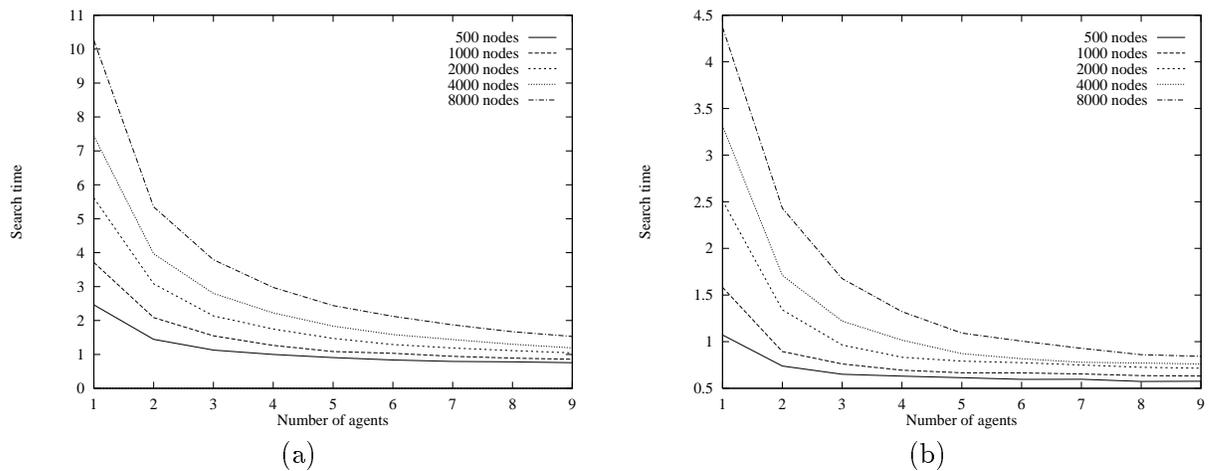

Figure 13: Time consumption as a function of the number of agents for various sizes of: (a) Sparse Delaunay graphs, and (b) dense Delaunay graphs.

Figures 13(a), (b) present the costs of the time-efficient algorithm as a function of the number agents for various sizes of sparse and dense Delaunay graphs, respectively. The results confirm the same tendency that was observed for regular Delaunay graphs (see

Figure 9), namely that as the number of agents grows, the overall cost converges to the length of the optimal path.